\documentclass[runningheads]{llncs}
\usepackage[T1]{fontenc}
\usepackage{microtype}
\usepackage{graphicx}
\usepackage{booktabs}
\usepackage{multirow}
\usepackage{amsmath,amssymb}
\usepackage{placeins}
\usepackage{marvosym}
\usepackage{xcolor}
\usepackage{xurl}
\usepackage[hyperfootnotes=false,breaklinks=true,colorlinks=true,linkcolor=blue,citecolor=blue,urlcolor=blue]{hyperref}
\hypersetup{
  pdftitle={Task-Performance Routing for Multi-Teacher Distillation in CT Medical Image Segmentation},
  pdfauthor={Jiaye Yang, Yao Liu, Die Dai, Hanwen Zhang, Peiyuan Jiang, Qiao Liu, Yutong Xie, Peng Wang},
  pdfsubject={MICCAI 2026; author-prepared version},
  pdfkeywords={multi-teacher knowledge distillation, CT segmentation, foundation models, task-performance routing}
}

\begin{document}
\raggedbottom
\title{Task-Performance Routing for Multi-Teacher Distillation in CT Medical Image Segmentation}
\titlerunning{Task-Performance Routing for CT Segmentation}
\author{Jiaye Yang\inst{1} \and
Yao Liu\inst{1}\textsuperscript{(\Letter)} \and
Die Dai\inst{1} \and
Hanwen Zhang\inst{1} \and
Peiyuan Jiang\inst{1} \and
Qiao Liu\inst{1} \and
Yutong Xie\inst{2} \and
Peng Wang\inst{1}
}
\authorrunning{J. Yang et al.}
\institute{University of Electronic Science and Technology of China, 
Chengdu, 611731, China\\
 \and
Mohamed bin Zayed University of Artificial Intelligence, 
Abu Dhabi, UAE \\ 
\email{liuyao@uestc.edu.cn}}
\maketitle
\begingroup
\renewcommand\thefootnote{}
\footnotetext{Author-prepared version of a paper accepted at MICCAI 2026.
Conference paper: \href{https://papers.miccai.org/miccai-2026/1031-Paper0973.html}{MICCAI Open Access}.}
\endgroup
\begin{abstract}
Pre-trained models offer complementary strengths in CT organ segmentation, but their relative accuracy varies across organs and spatial regions. Multi-teacher knowledge distillation can combine these strengths in a single student, yet global or instance-level teacher weights overlook variation within a 3D volume. We propose Task-Performance Routing (TPR), which uses region-wise segmentation errors to determine teacher weights during training without learned routing parameters. In hard regions, TPR favors teachers with lower prediction errors; in easy regions, it uses uniform weights to combine their knowledge. A hierarchical spatial partition provides fine-grained routing in shallow encoder stages and coarse-grained routing in deep stages. Experiments on BTCV, MSD-Liver, and MSD-Pancreas with two teacher pairs and two backbone architectures show higher Dice scores than the individual teachers, adaptive ensembles, and evaluated multi-teacher distillation baselines. On MSD-Pancreas, TPR achieves Dice scores of 65.40\% with SwinUNETR and 68.25\% with U-Net, compared with 64.38\% and 67.45\% for the corresponding adaptive ensembles. Inference uses the student alone, with no routing or teacher overhead. Code is available at \url{https://github.com/EdvinCecilia/TPR_MultiTeacher}.

\end{abstract}

\section{Introduction}

\begin{figure}[t]
\centering
\includegraphics[width=0.98\textwidth]{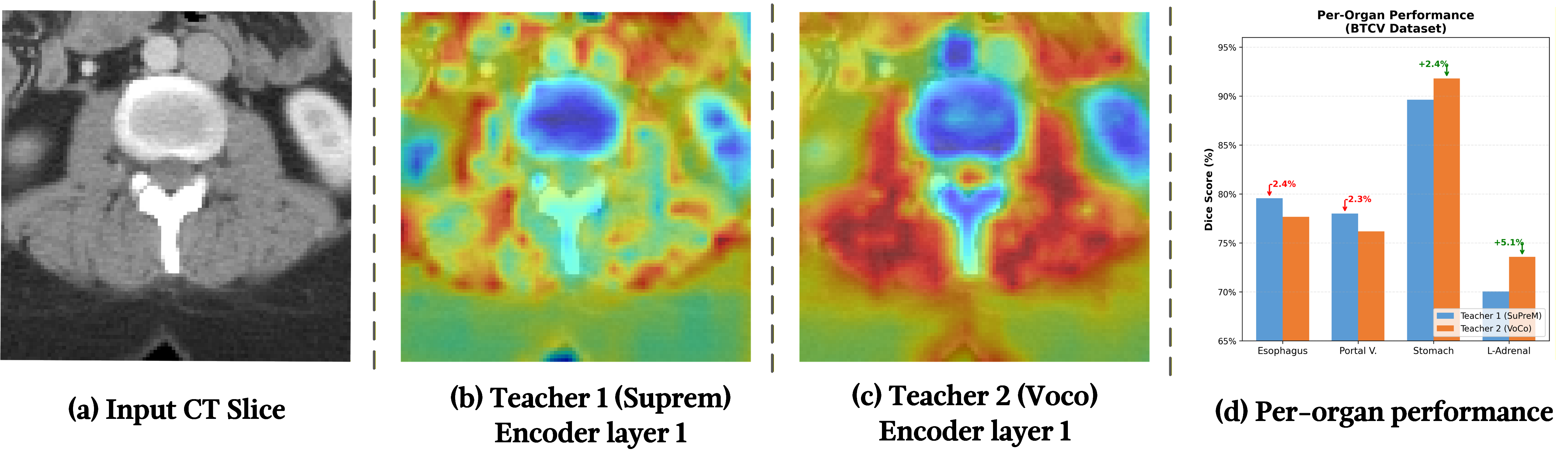}
\caption{Motivation on BTCV: Complementary performance of two teachers. \textbf{(a)} Input CT slice. \textbf{(b--c)} Encoder feature activation maps from SuPreM and VoCo reveal distinct spatial focus: SuPreM activates strongly around small tubular structures, whereas VoCo responds more broadly along large-organ boundaries. \textbf{(d)} Per-organ Dice scores confirm organ-wise complementarity: SuPreM (blue) excels at the esophagus and portal vein, whereas VoCo (orange) is stronger for the stomach and left adrenal gland.}
\label{fig:motivation}
\end{figure}

Large-scale pre-trained models have significantly advanced CT organ segmentation, yet different pre-training paradigms (e.g., SuPreM~\cite{li2024well}, VoCo~\cite{wu2024voco}, and CLIP-Driven~\cite{liu2023clip}) yield complementary representations, excelling in different organs and spatial regions due to their distinct training objectives and data distributions. 
We analyze SuPreM and VoCo pre-training representations on multi-organ segmentation data BTCV~\cite{landman2015miccai}. Despite similar overall Dice scores, they exhibit distinct spatial activation patterns and complementary organ-level performance (Fig.~\ref{fig:motivation}), 
suggesting that spatially adaptive routing can exploit these differences more effectively than global weighting schemes.

Multi-teacher knowledge distillation (MTKD)~\cite{hinton2015distilling,gou2021knowledge} offers a natural way to merge these strengths into a single student without inference overhead. 
Existing MTKD methods have evolved from uniform weighting~\cite{hossain2025enhancing} to instance-level adaptive schemes based on latent representations~\cite{liu2020adaptive,zagoruyko2017paying} or gradient-space optimization~\cite{du2020agree}, as well as multi-scale self-distillation strategies~\cite{jia2024mssd}. In medical imaging, MTKD has been primarily explored for cross-modal distillation~\cite{ahmad2024multi,hu2020knowledge}, while intra-modality settings remain underexplored~\cite{wang2024progressively}. Furthermore, prior work typically trains teachers from scratch rather than distilling from independently pre-trained foundation models---a setting in which the goal shifts from compression to knowledge fusion~\cite{ma2024segment}. These weighting strategies do not directly account for region-wise spatial heterogeneity within 3D CT volumes, where tissue contrast, boundary complexity, and organ morphology vary across spatial locations within the same sample.

To address this gap, we propose Task-Performance Routing (TPR), which dynamically assigns region-wise teacher weights based on segmentation errors during training, favoring lower-error teachers in hard regions while encouraging complementary fusion in easy regions. This performance-driven routing operates at hierarchical spatial granularity, with fine-grained routing in shallow layers for boundary details and coarse-grained routing in deep layers for semantic coherence, requiring no learned routing parameters while naturally adapting to the spatial heterogeneity of CT volumes.
Experiments on BTCV, MSD-Liver, and MSD-Pancreas with different teacher pairs show that TPR consistently outperforms individual teachers, adaptive ensemble methods, and existing MTKD baselines. Our main contributions are:
\begin{itemize}
\item \textbf{Task-performance routing:} region-wise teacher selection driven by prediction error with hierarchical spatial granularity, requiring no extra routing parameters.
\item \textbf{Complementarity-aware distillation:} effective knowledge fusion from independently pre-trained foundation models that surpasses both individual teachers and ensemble baselines without inference overhead.
\end{itemize}

\begin{figure}[t]
\centering
\includegraphics[width=\textwidth]{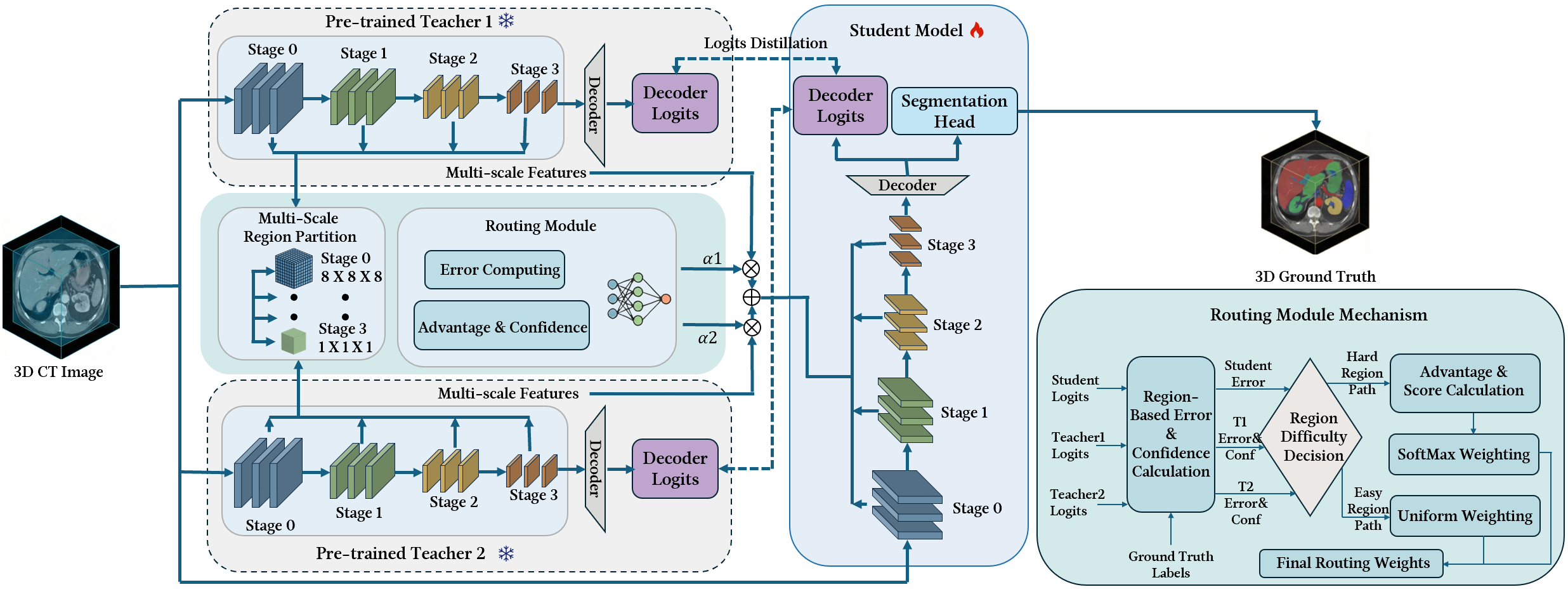}
\caption{Task-Performance Routing (TPR). Frozen teachers provide encoder features and predictions. During training, region-wise errors guide hierarchical teacher weighting for student distillation. At inference, only the student is used.}
\label{fig:framework}
\end{figure}

\section{Method}

Figure~\ref{fig:framework} provides an overview of TPR, illustrating the hierarchical routing mechanism and the feature alignment process across encoder stages.

\subsection{Overview and Problem Formulation}

Given an input CT volume $\mathbf{x} \in \mathbb{R}^{H\times W\times D}$ and its ground truth segmentation mask $\mathbf{y} \in \{0,1,...,C\}^{H\times W\times D}$ with $C$ organ classes, we train a student $\mathcal{S}_\theta$ using $M$ frozen teacher models $\{\mathcal{T}_m\}_{m=1}^{M}$. Each teacher is independently pre-trained under a different paradigm. Our goal is to enable the student to surpass the performance of individual teachers by effectively combining their complementary strengths across different spatial regions.

Both teachers and student share an identical encoder-decoder architecture: SwinUNETR~\cite{hatamizadeh2021swin} for VoCo and SuPreM, and U-Net~\cite{ronneberger2015u} for SuPreM and CLIP-Driven. They extract multi-scale features at $L$ stages. For the student, we denote the encoder features at stage $l$ as $\mathbf{f}_s^l \in \mathbb{R}^{C_l \times H_l \times W_l \times D_l}$ and the final output logits as $\mathbf{L}_s \in \mathbb{R}^{(C+1) \times H \times W \times D}$. Similarly, teacher $m$ produces features $\mathbf{f}_{t_m}^l$ and logits $\mathbf{L}_{t_m}$. All teacher models are completely frozen during training, requiring no gradient computation. We propose TPR, which computes adaptive routing weights $\mathbf{w}_m^{l,r}$ for teacher $m$ at stage $l$ and region $r$ based on actual segmentation performance, in contrast to traditional MTKD methods that rely on spatially uniform or globally adaptive weights.

\subsection{Task-Performance Router}

The TPR dynamically determines teacher weights based on actual prediction errors, comprising three components: hierarchical region division, routing weight generation, and hard/easy region adaptation.

\subsubsection{Hierarchical Region Division}

Medical image segmentation involves features at different abstraction levels---shallow layers capture fine-grained boundary details while deep layers encode global semantic context. To match routing granularity with feature hierarchy, we apply fixed 3D grid division at each encoder stage $l$.

Given a feature map $\mathbf{f}^l \in \mathbb{R}^{C_l \times H_l \times W_l \times D_l}$ at stage $l$, we divide it into $N_l$ non-overlapping 3D grid regions. Here $R_l$ denotes the number of grid divisions along the three spatial axes, rather than the voxel dimensions of an individual region. The grid resolution decreases from shallow to deep layers:

\begin{equation}
R_l = \begin{cases}
8 \times 8 \times 8, & l = 0 \text{ (shallow)} \\
4 \times 4 \times 4, & l = 1 \\
2 \times 2 \times 2, & l = 2 \\
1 \times 1 \times 1, & l = 3 \text{ (deep)}
\end{cases}
\end{equation}

Thus, $8\times8\times8$ specifies 512 regions, whereas $1\times1\times1$ specifies one global region. This hierarchical design reflects the intuition that shallow layers require fine-grained spatial routing to handle local variations in tissue contrast and boundary clarity, while deep layers benefit from coarse or global routing to maintain semantic coherence across organs.

\subsubsection{Routing Weight Generation}
We first compute a voxel-wise cross-entropy error map from the final logits 
of the student and each frozen teacher, then pool this map using the 
stage-specific 3D grid to obtain region-level errors $e_s^r$ and 
$e_{t_m}^r$ at each encoder stage. Regions with larger student 
error are treated as hard regions ($e_s^r > \tau_{\text{hard}}$, where 
$\tau_{\text{hard}}$ is a threshold hyperparameter), motivated by the 
observation that hard examples contribute disproportionately to 
learning~\cite{lin2017focal}. This hard/easy region distinction also 
draws on curriculum learning principles~\cite{bengio2009curriculum}, 
where harder examples receive more focused supervision. In hard regions, 
we favor teachers with lower errors using a softmax over each teacher's relative 
advantage; in easy regions, we apply uniform weights to encourage the 
student to absorb complementary knowledge from all teachers rather than 
overfitting to a single one:

\begin{equation}
w_m^{l,r} = \begin{cases}
\frac{\exp\left(\frac{e_s^r - e_{t_m}^r}{\tau}\right)}{\sum_{m'=1}^{M}\exp\left(\frac{e_s^r - e_{t_{m'}}^r}{\tau}\right)}, & e_s^r > \tau_{\text{hard}} \\
\frac{1}{M}, & \text{otherwise}
\end{cases}
\end{equation}

For a fixed region, $e_s^r$ is shared by all teachers and therefore cancels in the softmax normalization. The hard-region weights are equivalently $\operatorname{softmax}_m(-e_{t_m}^r/\tau)$; the student error determines which routing branch is used. Routing uses ground truth during training. At inference, only the student is evaluated.

\subsection{Multi-Component Loss Function}

The total training objective combines four terms:
\begin{equation}
\mathcal{L}_{\text{total}} = \lambda_{\text{seg}} \mathcal{L}_{\text{seg}} + \lambda_{\text{align}} \mathcal{L}_{\text{align}} + \lambda_{\text{logits}} \mathcal{L}_{\text{logits}} + \lambda_{\text{balance}} \mathcal{L}_{\text{balance}}
\end{equation}

\noindent\textbf{Segmentation loss} provides direct supervision from ground truth using a combination of Dice and cross-entropy losses: $\mathcal{L}_{\text{seg}} = \mathcal{L}_{\text{Dice}}(\mathbf{L}_s, \mathbf{y}) + \mathcal{L}_{\text{CE}}(\mathbf{L}_s, \mathbf{y})$.

\noindent\textbf{Feature alignment loss} aligns student encoder features with teacher features using region-adaptive routing weights across all $L=4$ encoder stages, from fine-grained shallow features to coarse semantic deep features. Here $\bar{\mathbf{f}}^{l,r}$ denotes the spatially averaged feature vector within region $r$ at stage $l$:
\begin{equation}
\mathcal{L}_{\text{align}} = \sum_{l=0}^{L-1} \frac{1}{N_l} \sum_{r=1}^{N_l} \sum_{m=1}^{M} w_m^{l,r} \left\| \bar{\mathbf{f}}_s^{l,r} - \bar{\mathbf{f}}_{t_m}^{l,r} \right\|_2^2
\end{equation}

\noindent\textbf{Logits distillation loss} transfers knowledge from teachers' final predictions via KL divergence~\cite{park2019relational} on region-averaged logits, where $\bar{\mathbf{L}}^r$ denotes the spatially averaged logit vector within region $r$:
\begin{equation}
\mathcal{L}_{\text{logits}} = \frac{1}{N_{\text{out}}} \sum_{r=1}^{N_{\text{out}}} \sum_{m=1}^{M} w_m^{L,r} \text{KL}\left(p_s(\bar{\mathbf{L}}_s^r) \| p_{t_m}(\bar{\mathbf{L}}_{t_m}^r)\right)
\end{equation}

\noindent\textbf{Load balance regularization} prevents routing collapse, 
inspired by load balancing in mixture-of-experts 
models~\cite{shazeer2017outrageously}. We maximize the entropy of average routing weights $\bar{\mathbf{w}}$:
\begin{equation}
\mathcal{L}_{\text{balance}} = \exp\left(-\frac{\mathcal{H}(\bar{\mathbf{w}})}{\tau_b}\right), \quad \bar{w}_m = \frac{1}{N_{\text{total}}} \sum_{l,r} w_m^{l,r}
\end{equation}
where $\mathcal{H}(\bar{\mathbf{w}}) = -\sum_{m=1}^M \bar{w}_m \log \bar{w}_m$ and $\tau_b=0.4$. We set $\lambda_{\text{seg}}=1.0$, $\lambda_{\text{align}}=0.2$, $\lambda_{\text{logits}}=0.2$, $\lambda_{\text{balance}}=0.1$.

\section{Experiments}

\subsection{Datasets and Evaluation Metrics}

We evaluate TPR on BTCV~\cite{landman2015miccai} and two MSD tasks~\cite{simpson2019large} (Liver and Pancreas). For BTCV, we follow the commonly-used split (24 training / 6 testing)~\cite{hatamizadeh2021swin}. For MSD, we use the official training set provided by the challenge and create a held-out test split (80\% for training and 20\% for testing, random seed 42) within it for all comparisons. All volumes are resampled to $1.5 \times 1.5 \times 2.0$ mm and intensities are clipped to $[-175,250]$ HU. We report mean Dice (\%, higher is better) and mean HD95 (mm, lower is better) over all foreground classes.

\subsection{Teacher Models and Experimental Settings}
We use two teacher pairs: (VoCo~\cite{wu2024voco}, SuPreM~\cite{li2024well}) with SwinUNETR and (SuPreM~\cite{li2024well}, CLIP-Driven~\cite{liu2023clip}) with U-Net; the student shares the same architecture as its teachers to enable feature alignment. All models use a patch size of $96\times96\times96$. Since teacher models only release encoder weights, we train teacher decoders from scratch for 1200 epochs with frozen encoders. Students are trained for 1200 epochs using AdamW~\cite{loshchilovdecoupled} (lr=$3\times10^{-4}$, weight decay=$10^{-5}$, 20 warmup epochs), with $\tau_{\text{hard}}=0.5$ and $\tau=0.7$. Data augmentation includes random flipping and rotation (p=0.2). All experiments use 4 NVIDIA RTX 4090 GPUs. The adaptive ensemble baseline applies softmax-normalized weights based on each teacher's validation Dice score.

\subsection{Comparison Study}

As shown in Table~\ref{tab:main_results}, TPR consistently outperforms both individual teachers, adaptive ensemble, and all MTKD baselines across three datasets and both teacher pair settings. Notably, TPR surpasses even the adaptive ensemble method, demonstrating that our distillation approach achieves superior knowledge fusion rather than simply aggregating teacher outputs.

\begin{table}[!htbp]
\centering
\caption{Comparison of the TPR student with individual teachers and multi-teacher distillation baselines on three datasets. T1/T2 denote VoCo/SuPreM for SwinUNETR and SuPreM/CLIP-Driven for U-Net. We report Dice (\%, higher is better) and HD95 (mm, lower is better). Best results are \textbf{bold}, second best \underline{underlined}.}
\label{tab:main_results}
\footnotesize
\setlength{\tabcolsep}{4pt}
\renewcommand{\arraystretch}{1.1}
\resizebox{\textwidth}{!}{%
\begin{tabular}{@{}llcccccc@{}}
\toprule
& & \multicolumn{3}{c}{\textbf{SwinUNETR-based}} & \multicolumn{3}{c}{\textbf{U-Net-based}} \\
& & \multicolumn{3}{c}{\textit{(VoCo + SuPreM)}} & \multicolumn{3}{c}{\textit{(SuPreM + CLIP)}} \\
\cmidrule(lr){3-5} \cmidrule(lr){6-8}
\textbf{Method} & \textbf{Metric} & BTCV & Liver & Panc. & BTCV & Liver & Panc. \\
\midrule
\multirow{2}{*}{T1} 
  & Dice (\%)  & 85.09 & 81.42 & 62.28 & 83.40 & 82.20 & 65.60 \\
  & HD95  & 2.57   & 14.65  & \underline{15.33}  & 2.68   & 13.95  & 16.12  \\[2pt]
\multirow{2}{*}{T2} 
  & Dice (\%)  & 84.50 & 81.79 & 63.72 & 84.20 & 83.00 & 67.37 \\
  & HD95  & 2.71   & 13.82  & 17.99  & \underline{2.55}   & 13.45  & \underline{15.68}  \\
\midrule
\multirow{2}{*}{Ensemble (adaptive)} 
  & Dice (\%)  & 85.15 & \underline{82.28} & 64.38 & 84.68 & 83.78 & 67.45 \\
  & HD95  & \underline{2.50}   & 13.48  & 15.62  & 2.62   & 13.48  & 15.92  \\[2pt]
\multirow{2}{*}{AE-KD~\cite{du2020agree} {\tiny(NIPS'20)}} 
  & Dice (\%)  & 85.10 & 81.85 & 63.58 & 84.38 & 83.22 & 66.72 \\
  & HD95  & 2.65   & 13.88  & 16.52  & 2.68   & 13.72  & 16.05  \\[2pt]
\multirow{2}{*}{MMKD~\cite{zhang2023adaptive} {\tiny(ICME'23)}} 
  & Dice (\%)  & \underline{85.25} & 81.98 & 63.89 & 84.52 & 83.58 & 67.12 \\
  & HD95  & 2.58   & 13.65  & 16.18  & 2.64   & 13.55  & 15.98  \\[2pt]
\multirow{2}{*}{MTKD-RL~\cite{yang2025multi} {\tiny(AAAI'25)}} 
  & Dice (\%)  & 85.23 & 82.15 & \underline{64.52} & \underline{84.75} & \underline{84.08} & \underline{67.58} \\
  & HD95  & 2.54   & \underline{13.38}  & 15.78  & 2.58   & \underline{13.32}  & 15.72  \\
\midrule
\multirow{2}{*}{\textbf{TPR (Ours)}} 
  & Dice (\%)  & \textbf{85.45} & \textbf{82.80} & \textbf{65.40} & \textbf{85.20} & \textbf{84.70} & \textbf{68.25} \\
  & HD95  & \textbf{2.40}   & \textbf{13.08}  & \textbf{15.24}  & \textbf{2.50}   & \textbf{13.15}  & \textbf{15.45}  \\
\bottomrule
\end{tabular}%
}
\end{table}

\subsection{Ablation Studies}
We conduct ablations in the SwinUNETR-based setting on MSD-Pancreas to validate the key design choices of TPR. All variants are trained for 1200 epochs under identical settings.

\begin{table}[!htbp]
\centering
\caption{Component ablation on MSD-Pancreas (SwinUNETR-based setting).}
\label{tab:ablation_compact}
\begin{tabular}{l|c}
\hline
Variant & Mean Dice (\%) \\
\hline
Teacher 1 (T1) & 62.3 \\
Teacher 2 (T2) & 63.7 \\
\hline
Baseline (uniform multi-teacher KD) & 63.9 \\
w/o global performance weighting & 64.3 \\
w/o spatial routing (hard/easy split) & 64.8 \\
w/o hierarchical granularity & 65.1 \\
\hline
\textbf{TPR (full)} & \textbf{65.4} \\
\hline
\end{tabular}
\end{table}

\begin{table}[!htbp]
\centering
\caption{Routing granularity strategies on MSD-Pancreas.}
\label{tab:ablation_granularity}
\setlength{\tabcolsep}{8pt}
\renewcommand{\arraystretch}{1.05}
\begin{tabular}{l|c}
\hline
Routing Strategy & Mean Dice (\%) \\
\hline
All fine-grained ($8\times8\times8$ at all stages) & 64.6 \\
All coarse-grained ($1\times1\times1$ at all stages) & 64.2 \\
Uniform granularity ($4\times4\times4$ at all stages) & 64.9 \\
\textbf{Hierarchical (ours)} & \textbf{65.4} \\
\hline
\end{tabular}
\end{table}

\begin{table}[!htbp]
\centering
\caption{Loss component ablation on MSD-Pancreas (SwinUNETR-based, 1200 epochs).}
\label{tab:ablation_remove}
\setlength{\tabcolsep}{6pt}
\renewcommand{\arraystretch}{1.05}
\begin{tabular}{c|c|c|c}
\hline
$\mathcal{L}_{align}$ & $\mathcal{L}_{logits}$ & $\mathcal{L}_{balance}$ & Mean Dice (\%)\\
\hline
$\checkmark$ & -- & $\checkmark$ & 64.52 \\
-- & $\checkmark$ & $\checkmark$ & 64.78 \\
$\checkmark$ & $\checkmark$ & -- & 65.11 \\
$\checkmark$ & $\checkmark$ & $\checkmark$ & \textbf{65.40} \\
\hline
\end{tabular}
\end{table}

As shown in Tables~\ref{tab:ablation_compact}--\ref{tab:ablation_remove}, each design choice contributes consistently and statistically significantly ($p < 0.05$, paired $t$-test).

Table~\ref{tab:ablation_compact} compares the routing components. Full TPR achieves 65.4\% Dice, improving on uniform multi-teacher distillation by 1.5 percentage points. Removing global performance weighting, spatial routing, or hierarchical granularity gives 64.3\%, 64.8\%, and 65.1\%, respectively. These comparisons support combining performance-based teacher weighting with spatial adaptation across encoder stages. $\tau_{\text{hard}}=0.5$ corresponds to regions exceeding the dataset mean cross-entropy error, and yields stable performance across $[0.4, 0.6]$.

Table~\ref{tab:ablation_granularity} isolates routing granularity strategies. Fine-grained routing at deep layers introduces spatial noise that disrupts semantic coherence, while coarse routing at shallow layers fails to capture local boundary variations. Uniform granularity ($4\times4\times4$) falls short of hierarchical design by 0.5 percentage points.

Table~\ref{tab:ablation_remove} shows that all three loss components contribute to the full model. Removing $\mathcal{L}_{\text{align}}$, $\mathcal{L}_{\text{logits}}$, or $\mathcal{L}_{\text{balance}}$ reduces Dice by 0.62, 0.88, and 0.29 percentage points, respectively. The full objective combines feature-level alignment, output-level distillation, and balanced teacher utilization to achieve the highest Dice score among these variants.

\begin{figure}[t]
\centering
\includegraphics[width=\textwidth, height=10cm, keepaspectratio]{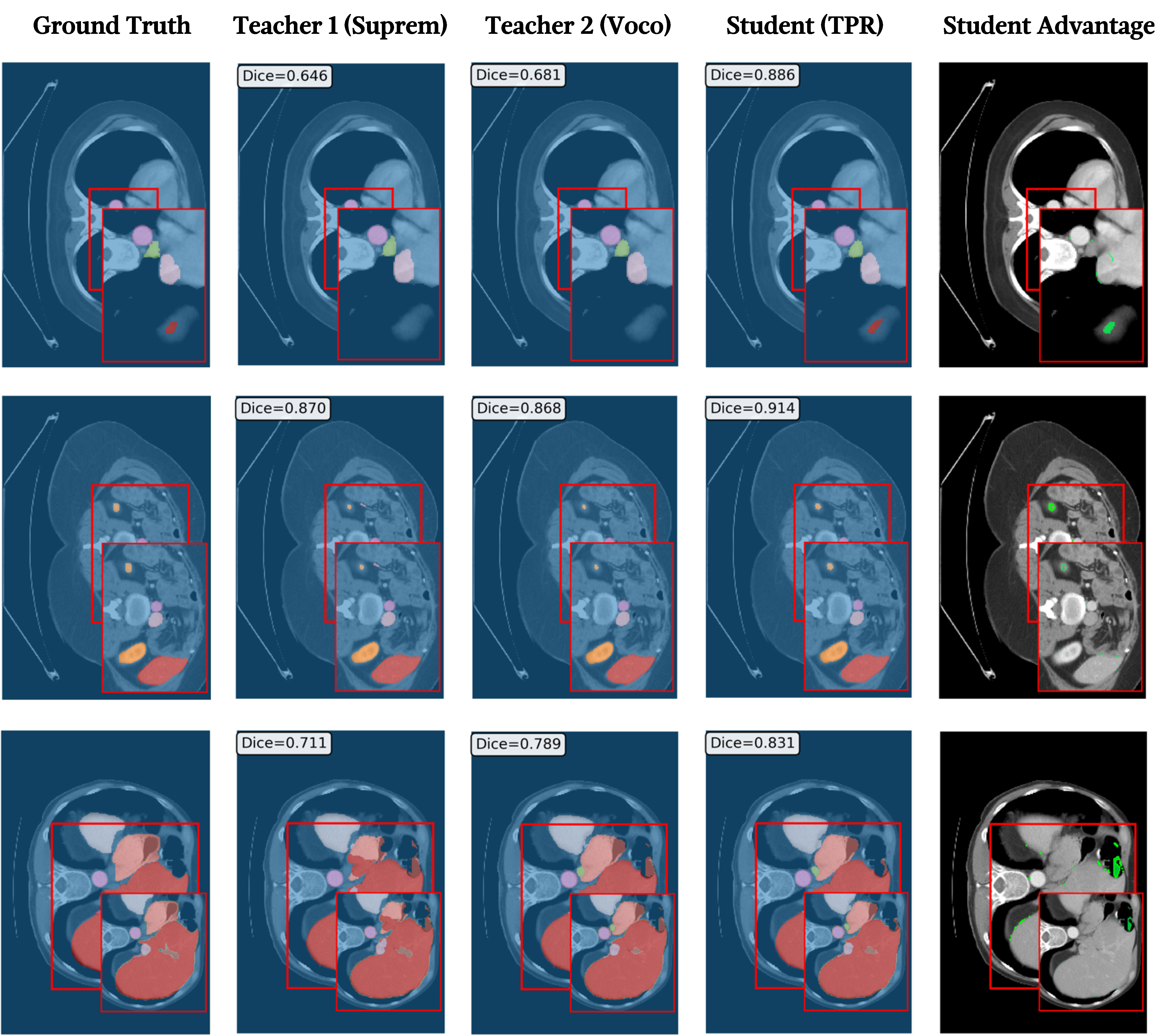} 
\caption{Qualitative comparison on BTCV. Columns show ground truth, SuPreM, VoCo, the TPR student, and the Student Advantage map. Red boxes highlight local details; green marks indicate regions where the student improves on both teachers. Teacher numbering follows this figure.}
\label{fig:qualitative}
\end{figure}

\FloatBarrier
\subsection{Qualitative Results}
Figure~\ref{fig:qualitative} presents representative CT slices from the BTCV dataset, comparing segmentation outputs of both teachers and our student against the ground truth. The rightmost column visualizes the \textit{Student Advantage} map, highlighting regions where the student outperforms both teachers simultaneously.

Teacher 1 (SuPreM) and Teacher 2 (VoCo) exhibit complementary failure patterns: SuPreM tends to over-segment neighboring organs, while VoCo frequently misses fine-grained boundaries in cluttered regions. Despite being distilled from these imperfect teachers, TPR produces boundaries that more closely align with the ground truth, as shown in the zoomed-in red box regions. The Student Advantage maps highlight local corrections beyond the predictions of either individual teacher.

\section{Conclusion}
We propose Task-Performance Routing (TPR) for multi-teacher distillation 
in CT segmentation, assigning region-wise teacher weights based on task 
error at hierarchical spatial granularity. TPR effectively exploits 
complementary strengths of different foundation models and yields 
consistent improvements over individual teachers. One limitation is that 
TPR requires ground truth labels for error-based routing, making it less 
suitable for semi-supervised or label-scarce settings; pseudo-label-based 
routing remains future work.

\bibliographystyle{splncs04}
\bibliography{ref}
\end{document}